\documentclass[pdflatex,sn-mathphys-num]{sn-jnl}% Math and Physical Sciences Numbered Reference Style
\usepackage{graphicx}%
\usepackage{multirow}%
\usepackage{amsmath,amssymb,amsfonts}%
\usepackage{amsthm}%
\usepackage{mathrsfs}%
\usepackage[title]{appendix}%
\usepackage{xcolor}%
\usepackage{textcomp}%
\usepackage{manyfoot}%
\usepackage{booktabs}%
\usepackage{algorithm}%
\usepackage{algorithmicx}%
\usepackage{algpseudocode}%
\usepackage{listings}%
\usepackage{float}
\theoremstyle{thmstyleone}%
\theoremstyle{thmstyletwo}%

\theoremstyle{thmstylethree}%

\begin{document}

\setcounter{secnumdepth}{3}
\setcounter{tocdepth}{3}

\title[Article Title]{Machine Learning to Predict Digital Frustration from Clickstream Data}


\author*[1]{\fnm{Jibin} \sur{Joseph}}\email{jibinjoseph@utexas.edu}

\affil*[1]{\orgdiv{Department of Computer Science},
\orgname{The University of Texas at Austin},
\orgaddress{\city{Austin}, \state{Texas}, \country{USA}}}


\abstract{Many businesses depend on their mobile apps and websites, so user frustration while trying to complete a task on these channels can cause lost sales and complaints. In this research, I use clickstream data from a real e-commerce site to predict whether a session is frustrated or not. Frustration is defined using certain rules based on rage bursts, back-and-forth navigation (U-turns), cart churn, search struggle, and long wandering sessions, and applies these rules to 5.4 million raw clickstream events (304,881 sessions). From each session, I build tabular features and train standard classifier models. I also use the full event sequence to train a discriminative LSTM classifier. XGBoost reaches about 90\% accuracy, ROC AUC of 0.9579, while the LSTM performs best with about 91\% accuracy and a ROC AUC of 0.9705. Finally, the research shows that with only the first 20-30 interactions, the LSTM already predicts frustration reliably.}




\maketitle

\section{Introduction}\label{sec1}

Modern businesses increasingly operate through their digital channels, such as mobile applications or websites. Retail e-commerce has grown significantly since 2010 to one-fifth of total retail sales today \cite{1} \cite{2}. For banks and financial institutions, recent studies show that around three-quarters of millennials prefer digital channels to manage their bank accounts \cite{3}. The number of online banking users reached 3.6 billion in 2025, projected to reach \$107 billion in revenue by 2030 \cite{4}. The popularity of affordable smartphones, internet connectivity, and digital payments accelerated this trend. More than 85\% of households own a smartphone in India, and 99\% of young people use digital payment systems based on the Unified Payments Interface (UPI) \cite{5}.

So, each business's digital experience became more important. A smooth, fast, and positive digital experience is associated with higher customer satisfaction and loyalty that can drive growth \cite{6}. On the other hand, poor digital experience has its consequences like higher customer churn, negative feedback, and reputation loss. A study on cart abandonment rate in the e-commerce sector showed that around 71\% of users abandon their purchase, and a large fraction among them do so due to poor digital experience \cite{7}. A performance study by Cloudflare \cite{8} showed that increased page load time leads to a significant drop in conversions, which leads to large revenue losses.

In this research, I focus on building a Machine Learning framework that can identify digital frustration from clickstream data, the record of a user's journey through a digital channel, including pages visited and actions taken. Here, my objective is to
\begin{enumerate}
\item Define user frustration on digital channels by combining interaction signals such as rage clicks, repeated back-and-forth navigation, and long dwell times of pages.

\item Create a practical labeling strategy.
Engineer a set of features from sessionized clickstream data, including behavioural, statistical, and sequential features.

\item Develop and compare classical machine learning models and deep learning based sequence models for predicting whether a user's session is likely to be frustrated or not.

\item How early in the user's session can a reliable prediction be made?

The main contribution of this research is a reproducible pipeline that converts the raw user interaction data on digital channels to frustration labels, creates tabular features for classical models, sequence modeling, and compares model performance for frustration detection, and an analysis on early window prediction.
\end{enumerate}

\section{Background and Related Work}\label{sec2}

\subsection{Industry view of frustration signals}\label{subsec2}

There are many digital experience analytics and optimization platforms in the industry that continuously capture user interaction data from digital channels and identify frustration signals.

Fractal’s AIDE uses its sensorize method with AI to extract insights from user interactions. It detects deviations in user behaviour and identifies their impact on the user’s digital experience \cite{9}. FullStory tracks specific frustration signals such as rage clicks, dead clicks, error clicks, and thrashed cursor movements and uses this information to segment digital sessions and identify likely problems \cite{10}. ContentSquare uses a frustration score that quantifies the digital frustration to indicate how difficult the user experience is at the page level or session level \cite{11}. Glassbox provides auto detection of around thirty user behaviours, from rage clicks, slow load time, JavaScript errors, to indicate user struggle \cite{12}. Quantum Metric also captures behavior and frustration signals, technical metrics, and provides real-time digital experience alerts \cite{13}.

All these commercial platforms show that digital frustration can be inferred from capturing a large amount of user interaction data or clickstream data, and by analyzing the patterns in it with the help of machine learning.

\subsection{Literature on clickstream modeling}\label{subsec2}

The academic work on clickstream data ranges from exploratory data analysis, user intent prediction, sequence representation learning, and frustration detection.  One of the early works, such as Wei et al (2012) \cite{14}, proposed a visual analytics system that clusters user sessions into different behavioural groups and lets analysts make inferences out of those patterns. Wang et al. (2017) \cite{15} introduced 'clickstream user behaviour models' that represent user actions as sequences, build similarity graphs, and use clustering and classification to identify anomalous behaviours.

A substantial set of work applies machine learning to predict shopping or purchase intention from e-commerce clickstream data. Requena et al. (2020) \cite{16} use the Coveo e-commerce dataset \cite{17} and show that high-quality shopper intent prediction and useful early window intent prediction are possible with minimal browsing information using standard classifiers as well as LSTM models. Bakker (2020) \cite{18} compares different recurrent neural network architectures (Vanilla RNN, LSTM, GRU) on clickstream data and finds that LSTMs, when trained on full sequences, surpass baseline models for shopping-intent prediction. There are more works by Al-Tayeb (2024) \cite{20}, Yadav (2023) \cite{21}, Baati et al. (2020) \cite{22}, Anitha (2024) \cite{23}, and Le Tran Minh Nghia (2024) \cite{24} that experiment with a range of machine learning to predict whether users are going to complete a purchase or not.

Beyond prediction shopping or purchase intention, several studies look at early prediction and representation learning from behavioural sequences. Ulitzsch et al. (2022) \cite{19} propose a method that segments early window clickstream data from user interactions to predict eventual failure. Wang et al. (2021) \cite{25} introduce a user behaviour coding method that learns high-level embeddings of user behaviour sequences by maximizing mutual information between different views of the sequences. Ziming Wang et al. (2023) \cite{26} take a similar stand in 'Sequence As Genes' by considering user action sequences as gene-like units and design a modeling framework that detects fraud transactions in e-commerce. These generic sequence encoders can support a variety of behavior-prediction tasks.

Closer to my research on user frustration detection on digital channels,  Sanni (2024) \cite{27} conducts anomaly detection and user frustration prediction on mobile applications, combining behavioural and technical data using machine learning.  Rahmiati et al. (2022) \cite{28} analyze user behaviour on mobile banking and show how ease of use, habit, and value contribute to continued use of digital banking channels.

\section{Dataset and Problem Formulation}\label{sec2}

This research is based on the Coveo e-commerce clickstream dataset \cite{29}, which is made available to the AI community under a research friendly license. The dataset has 5.43M anonymized raw clickstream events from a fashion e-commerce website. Each row corresponds to a browsing event, and it has a hashed session identifier, event type, product action, time stamp, and hashed page URL.  Based on Requena et al. (2020) \cite{16}, this research sessionizes the raw clickstream data and symbolizes product actions to treat the frustration detection task as a session-level supervised classification problem.

\subsection{Symbolization and Sessionization}\label{subsec2}
For each event, I symbolize the product action by mapping it to a numerical symbol as per Table 1.  The raw clickstream data is then grouped by session id hash, maintaining the chronological order within each session using the server timestamp. This follows the same philosophy as Requena et al. (2020) \cite{16}, but the order of symbolization is slightly adjusted because I think the click action will happen before an add to cart action or purchase in a natural e-commerce purchase flow.

\begin{table}[h]
\caption{Symbolization}\label{tab1}%
\begin{tabular}{@{}llll@{}}
\toprule
Product Action & Description  & Symbol \\
\midrule
NaN (View)    & Any page load with no specific product action (NaN). Consider it as View   & 1 \\
Detail    & Product detail view   & 2 \\
Click    & Click on a search result   & 3  \\
Add    & Add to cart   & 4  \\
Remove    & Remove from cart   & 5  \\
Purchase    & Product purchase   & 6  \\
\botrule
\end{tabular}
\footnotetext{Product actions and corresponding symbols.}
\end{table}

\begin{table}[h]
\caption{Sessionization}\label{tab1}%
\begin{tabular}{@{}llll@{}}
\toprule
Session id hash & Symbol \\
\midrule
00007d15aeb741b3cdd873cb3933351d699cc320    & [1, 1, 2, 1, 2, 1, 2] \\
\botrule
\end{tabular}
\footnotetext{An example session with symbolized interaction sequence}
\end{table}

\subsection{Defining Frustration and Labeling}\label{subsec2}

Generally, digital frustration of a user can be defined as the inability to reach their goal, requiring excessive effort, such as showing rage clicks, backtracking (U-turns), or repeated failures \cite{31} \cite{32}. In this research, I define the following rules to identify frustrated user sessions.

\textbf{Rage burst}: Three or more consecutive events on the same page URL, with no more than two seconds between consecutive events, are considered as rage bursts \cite{30}.

\textbf{U-turns}: A U-turn occurs when a user navigates from page A to B and returns to A within two seconds, without any add to cart or purchase actions in between.

\textbf{Cart churn}: A session is considered to show cart churn if it contains both add to cart and remove from cart events, in any order, without a purchase.

\textbf{Search struggle}: Three or more search result clicks without any add to cart or purchase event are considered search struggle \cite{33}.

\textbf{Long wandering}: A session lasting more than 20 minutes, with at least five product detail views and no purchase, is considered long wandering.

If any of the above signals are present (rage bursts $> $ 0, U-turns $> $ 0, cart churn, search struggle, or long wander), I label the session as frustrated (label = 1); otherwise, non-frustrated (label = 0). By design, some of these rules decouple frustration from conversion. A session may still end in a purchase, but it gets the frustrated label if there are intense rage bursts or repeated U-turns before completion.

\begin{table}[h]
\centering

\begin{minipage}{0.48\linewidth}
\centering
\caption{Breakdown}\label{tab1-left}
\begin{tabular}{@{}llll@{}}
\toprule
Frustration Type  & Count \\
\midrule
Rage bursts   & 15442 \\
U turns   & 1753 \\
Cart churn       & 8434  \\
Search struggle       & 4829  \\
Long wander      & 47710  \\
\botrule
\end{tabular}
\footnotetext{Breakdown of the frustration types.}
\end{minipage}
\hfill
\begin{minipage}{0.48\linewidth}
\centering
\caption{Breakdown}\label{tab1-right}
\begin{tabular}{@{}llll@{}}
\toprule
Sessions  & Count \\
\midrule
Total sessions       & 443660  \\
Frustrated      & 57736  \\
Non frustrated  & 385924  \\
Sessions with single interaction  & 138656  \\
Sessions with more than 1000 interactions & 111  \\
\botrule
\end{tabular}
\footnotetext{Total sessions and other session groups.}
\end{minipage}

\end{table}

\subsection{Removing class information from page sequence to prevent short-cut learning}\label{subsec2}

Some of the frustration rules directly use conversion-related events such as add to cart, remove, and purchase.
Cart churn depends on the presence of add and remove, and the absence of a purchase event.
Search frustration depends on repeated search result clicks, with no add to cart and no purchase.
Long wander is when the session has many product detail views, but no purchase.

Because some frustration labeling rules depend on the absence of purchase events, the presence of the purchase symbol in the training data could encourage shortcut learning: A binary classifier model could trivially learn that "sessions with a purchase action are generally non-frustrated, although some may still show frustration" rather than learning complex patterns that cause frustration. This issue is severe for models such as an LSTM classifier because they could easily do shortcut learning. If the purchase symbol is present anywhere in the sequence, the LSTM could rely on that symbol to predict non-frustration without learning any complex interaction patterns that may present in the session. In this case, the model would appear to perform well even on unseen data, but it would not generalize to settings where we need to do early prediction with the first five or ten interactions in the sequence, where the purchase event is not directly observable.

To avoid this, right after performing the frustration labeling, I truncate each session at the first purchase event. Only the user interactions up to the first purchase are used for feature engineering and modeling. The labeling function I used will have access to the full user interactions, but the predictive models only see the interactions before the first purchase. This design will reduce shortcut learning and force both classical models and LSTM-based sequence models to learn complex behavioural patterns from the training data that can genuinely predict frustration rather than relying solely on the purchase symbol.

\section{Feature Engineering for Machine Learning}\label{sec2}

We treat the frustration prediction as a binary classification problem. Our sessionized data has sequences of user interaction and sequences of server timestamps as features at the moment. But a standard classifier model like Logistic Regression, SVM, Random Forest, or XGBoost cannot be directly trained on these sequences because they are variable in length. These models need tabular (structured data) features as predictors. Therefore, we need to summarize each interaction sequence with feature engineering methods that capture how the user behaved during the session. We preprocess the data and apply the following feature engineering methods to extract tabular features from sequential data

\subsection{Preprocessing}\label{subsec2}

After sessionizing the raw interactions (5,433,611 records), we get 443,660 individual sessions. As discussed in section 3.1, sessionization is the process of grouping raw interactions based on session id and ordering by timestamp. 
Then I apply the following preprocessing steps :

\begin{enumerate}

\item Trim the sequences by removing the interactions from the first purchase (explained in section 3.3)
\item There are 138,656 sessions in our data with only one interaction in the entire session. These are likely dropouts or accidental visits, and those will be removed.
\item There are a few sessions with a very large number of user interactions. To keep our computation manageable for this project, I’m dropping sessions with more than 1000 interactions, and there are only 111 sessions
in this group.

\end{enumerate}

After the initial length filtering, 304,893 sessions remain. An additional 12 sessions become shorter than two interactions after purchase truncation, leaving 304,881 sessions for modeling. I use the frustration labeling rules defined in Section 3.2 for labeling, and the labeled dataset contains 57,642 frustrated sessions (18.91\%) and 247,239 non-frustrated sessions.

\subsection{Sampling}\label{subsec2}
In this research, I'm performing the modeling on a balanced dataset with 50\% frustrated and non-frustrated sessions. A stratified sampling technique is used to create the balanced dataset. The dataset is split into frustrated and non-frustrated groups, and samples an equal number of training (70\%), test (15\%), and validation (15\%) data from both groups. For each split, now we have the following data.

\begin{enumerate}
\item Train set size: 80698 frustrated: 40349 non-frustrated: 40349
\item Test set size: 17294 frustrated: 8647 non-frustrated: 8647
\item Validation set size: 17292 frustrated: 8646 non-frustrated: 8646
\end{enumerate}

\subsection{Sequence-based features: n-grams and motifs}\label{subsec2}

First, I capture features from how symbols are arranged within each session. Here, I follow the “minimal information” philosophy from Requena et al. (2020) \cite{16}, who showed in the research paper that simple k-gram \cite{34} statistics and horizontal visibility graph motifs (HVGm) \cite{37}\cite{38} are very powerful for users’ purchase-intent prediction on the same Coveo e-commerce dataset \cite{29}.

\subsection{N-gram (k-gram) features}\label{subsec3}

An n-gram (or k-gram) is a short block of k consecutive symbols in the interaction sequence \cite{34}. For example, in a symbolized sequence like 1 2 2 3, the 1-grams are {1,2,2,3} and the 2-grams are {(1,2), (2,2), (2,3)}.
Based on this logic, for each session, we compute:
\begin{enumerate}
\item 1-gram features: Normalized frequencies P(s) of view, detail, search click, and add to cart events.

\item 2-gram features: Normalized frequencies P(s1,s2) for each pair of consecutive events.
\end{enumerate}

These features tell us how often a page appears in the user’s sequence and how often the user goes from view → detail or detail → add, which is already proven discriminative in the purchase intent prediction research paper \cite{16}.

Also, using n-grams as features is a standard feature engineering method in sequence/text classification. These sequences are represented by the counts or frequencies of n-grams and then used for training standard classifiers \cite{34}\cite{39}. My use of 1-grams and 2-grams for clickstream data follows this general approach, but applied to the symbolized digital interaction sequences instead of words.

\subsection{Horizontal visibility graph motifs (HVG motifs)}\label{subsec3}

N-grams are simple frequency counts and tell us which symbols immediately follow others, but they are not useful to capture more complex local patterns. To go beyond, we also use horizontal visibility graph motifs (HVGm), introduced by Iacovacci \& Lacasa (2016). \cite{37}\cite{38}. You can think of a motif as a subsequence of length p or a subgraph of p nodes.
Here, the idea is instead of converting the symbolized interaction sequences into a horizontal visibility graph and applying a sliding window algorithm \cite{35}\cite{36}, to use the simple inequality rule mentioned in the paper by Iacovacci \& Lacasa \cite{37}\cite{38} to build the motifs. In this research, I take the motif order p=4 based on the evidence in \cite{16}

For example (rule for motif Z2):
\begin{enumerate}
\item Take the 4 consecutive values in a sequence (x0, x1, x2, x3)
\item Check whether both conditions are satisfied:
$x1 < min(x0, x2)$, $x2 >= min(x0, x3)$
\item If both conditions are satisfied, increase the Z2 motif count by 1.
\item Take the next four consecutive values in the same sequence (x1, x2, x3, x4), if exist,  and repeat 2-3
\item At the end, convert the counts to probabilities
i.e., \[
Z2 = \frac{\text{count of motif } Z2 \text{ in the given sequence}}{\text{total number of motifs}}
\]
Where the total number of motifs  is the length of the sequence - 3 if the length is greater than 3

\end{enumerate}
For each session, we compute the Z = (Z1, Z2, Z3, Z4, Z5, Z6), where Zi is the probability of observing motif i.

The motif entropy HZ, which measures how diverse the motif profile is (Shannon's entropy, high entropy = many different navigation patterns, low entropy = very repetitive behavior).

In my work, I adopt the same family of 1-gram, 2-gram, and HVG motif features \cite{16}, but now with a frustration label instead of a purchase intent. So this research is also an experiment to test whether the patterns that help to classify buying behavior also help distinguish frustrated vs non-frustrated sessions.

\subsection{Cyclical features}\label{subsec3}

The second group of features captures the seasonality of the sessions. Many digital behaviors show strong time-of-day and day-of-week patterns, and machine learning models often benefit from cyclical patterns explicitly \cite{40}.

One way to add time information as features is to use integer variables such as “hour from 0 to 23” or “days from 0 to 6”. However, this treats hour 23 and hour 0 as far apart, but they are adjacent in time. To preserve the circular nature of clocks and calendars, a common trick is to map each cyclical variable onto the unit circle using sine and cosine \cite{40}.

Our dataset covers only a short period (from 8 December 2018 to 25 December 2018, which is the shifted time to anonymize the data), so we do not attempt to model longer seasonal effects (monthly or yearly). Instead, for each session, we take the timestamp of the first event from the sequence and derive:

\begin{enumerate}
\item 
Hour-of-day sine and cosine

\[
\text{hour\_sin} = \sin\!\left( 2\pi \frac{h}{24} \right), \qquad
\text{hour\_cos} = \cos\!\left( 2\pi \frac{h}{24} \right)
\]

\item 
Day-of-week sine and cosine

\[
\text{dow\_sin} = \sin\!\left( 2\pi \frac{d}{7} \right), \qquad
\text{dow\_cos} = \cos\!\left( 2\pi \frac{d}{7} \right)
\]
\end{enumerate}

\section{Modeling and Results}\label{sec2}

The digital frustration detection using machine learning is a binary classification problem. I chose three standard classifiers, Logistic regression, Random forest, and XGBoost, that take the engineered features (section 4.3 - 4.6) as predictors to predict frustration. Also, I use one sequence model, a discriminative LSTM classifier, that takes the vector embeddings of the interaction symbols as input and predicts the frustration.

\begin{enumerate}
\item \textbf{Yeo-johnson transformation}: Tree-based models are not sensitive to the scale of the numerical features, but still, skewness can have an effect. Some of the numerical features are right-skewed in the training data, and I need to transform the numerical features to reduce the skew. The cyclical features have negative, zero, and positive values in it, and one of the best transformation methods that can handle negative, zero, and positive values is the Yeo-Johnson transformation. I apply the Yeo-Johnson transformation \cite{44} on the training data and use the transformation parameters from the training data to transform the validation and test data. This prevents any data leakage during transformation, controls the variance, and reduces the skew of the numerical features. 

\item \textbf{Evaluation \& metrics}: For the model evaluation,  I chose the metrics accuracy, F1-score, ROC-AUC, and the classification report (precision, recall, F1 per class) on the unseen test set. While training the standard classifier models, the training vs validation curve is generated based on logloss to detect whether the model is overfitting or underfitting. For the LSTM classifier, I chose the loss function BCEWithLogitsLoss. It optimizes the binary cross-entropy between predicted labels and the ground truth labels.
\end{enumerate}

\subsection{Logistic Regression}\label{subsec2}

The logistic regression is the simplest model I train in the process of finding a baseline. The model gets about 84\% accuracy, F1-score 0.85, and ROC AUC 0.92 on the test set. It is good at finding frustrated sessions (class 1), with high recall (about 0.90), but it sometimes marks non-frustrated sessions as frustrated - recall 0.78 for class 0 based on the classification report (Table 5).

\begin{table}[h!]
\centering
\begin{minipage}{0.48\textwidth}
\centering
\caption{Logistic Regression Performance Metrics}
\begin{tabular}{lc}
\toprule
\textbf{Metric} & \textbf{Value} \\
\midrule
Accuracy & 0.8405 \\
F1-score & 0.8492 \\
ROC AUC & 0.9156 \\
\bottomrule
\end{tabular}
\end{minipage}
\hfill
\begin{minipage}{0.48\textwidth}
\centering
\caption{Classification report of Logistic Regression on the test set}
\begin{tabular}{lcccc}
\toprule
\textbf{Class} & \textbf{Precision} & \textbf{Recall} & \textbf{F1-score} & \textbf{Support} \\
\midrule
0 & 0.8853 & 0.7822 & 0.8306 & 8647 \\
1 & 0.8050 & 0.8987 & 0.8492 & 8647 \\
\midrule
Accuracy & & & 0.8405 & 17294 \\
Macro avg & 0.8451 & 0.8405 & 0.8399 & 17294 \\
Weighted avg & 0.8451 & 0.8405 & 0.8399 & 17294 \\
\bottomrule
\end{tabular}
\end{minipage}
\end{table}
\vspace{-15pt}

\subsection{Random Forest}\label{subsec2}
The Random Forest model performs better than logistic regression. It reaches about 89\% accuracy, F1-score 0.90, and ROC AUC 0.95 on the test set. It is strong for both classes, with good precision and recall for frustrated (class 1) and non-frustrated (class 0) sessions. The overall evaluation metric, classification report on the unseen test set are shown in Tables 7 and 8.

\begin{table}[h!]
\centering

\begin{minipage}{0.47\textwidth}
\centering
\caption{Random Forest Performance Metrics}
\begin{tabular}{lc}
\toprule
\textbf{Metric} & \textbf{Value} \\
\midrule
Accuracy & 0.8930 \\
F1-score & 0.8961 \\
ROC AUC & 0.9510 \\
\bottomrule
\end{tabular}
\end{minipage}
\hfill
\begin{minipage}{0.50\textwidth}
\centering
\caption{Classification report of Random Forest on the test set}
\begin{tabular}{lcccc}
\toprule
\textbf{Class} & \textbf{Precision} & \textbf{Recall} & \textbf{F1-score} & \textbf{Support} \\
\midrule
0 & 0.9181 & 0.8631 & 0.8897 & 8647 \\
1 & 0.8708 & 0.9230 & 0.8961 & 8647 \\
\midrule
Accuracy & & & 0.8930 & 17294 \\
Macro avg & 0.8944 & 0.8930 & 0.8929 & 17294 \\
Weighted avg & 0.8944 & 0.8930 & 0.8929 & 17294 \\
\bottomrule
\end{tabular}
\end{minipage}
\end{table}
\vspace{-15pt}

\subsection{XGBoost}\label{subsec2}
The XGBoost model gives the best results among the standard classifiers. The model gets about 90\% accuracy, F1-score 0.90, and ROC AUC 0.96 on the test set. It shows high precision and recall for both frustrated (class 1) and non-frustrated (class 0) sessions, with slightly higher recall for frustrated users (0.93). I applied further feature explainability on the XGBoost by creating the feature importance (Fig. 2) and SHAP (Fig. 3) to showcase what features are important and their effect on each class.

 Let's consider the results of XGBoost as our baseline for the digital frustration classification task and try to beat its performance using sequence modeling based on deep learning (section 5.4).

\begin{figure}[H]
    \centering
    \includegraphics[width=0.8\textwidth]{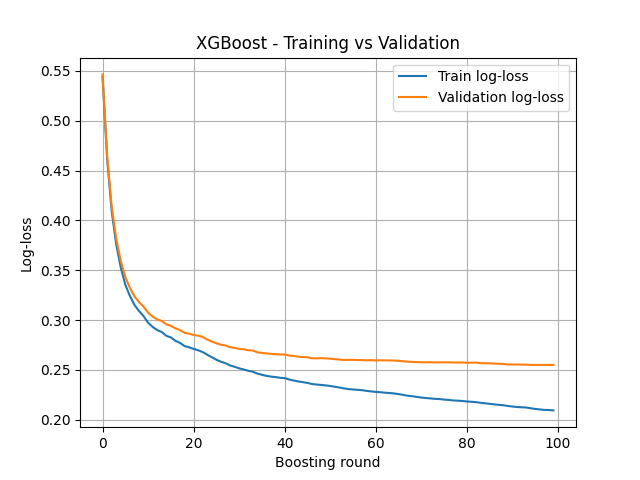}
    \centering
    \caption{The training vs validation curve shows that the XGBoost model trains very well}
    \label{fig:chart}
\end{figure}

\begin{table}[h!]
\centering

\begin{minipage}{0.47\textwidth}
\centering
\caption{XGBoost Performance Metrics}
\begin{tabular}{lc}
\toprule
\textbf{Metric} & \textbf{Value} \\
\midrule
Accuracy & 0.8995 \\
F1-score & 0.9025 \\
ROC AUC & 0.9579 \\
\bottomrule
\end{tabular}
\end{minipage}
\hfill
\begin{minipage}{0.50\textwidth}
\centering
\caption{Classification report of XGBoost on the test set}
\begin{tabular}{lcccc}
\toprule
\textbf{Class} & \textbf{Precision} & \textbf{Recall} & \textbf{F1-score} & \textbf{Support} \\
\midrule
0 & 0.9261 & 0.8683 & 0.8963 & 8647 \\
1 & 0.8760 & 0.9307 & 0.9025 & 8647 \\
\midrule
Accuracy & & & 0.8995 & 17294 \\
Macro avg & 0.9011 & 0.8995 & 0.8994 & 17294 \\
Weighted avg & 0.9011 & 0.8995 & 0.8994 & 17294 \\
\bottomrule
\end{tabular}
\end{minipage}
\end{table}
\vspace{-15pt}

\begin{figure}[H]
    \centering
    \includegraphics[width=0.8\textwidth]{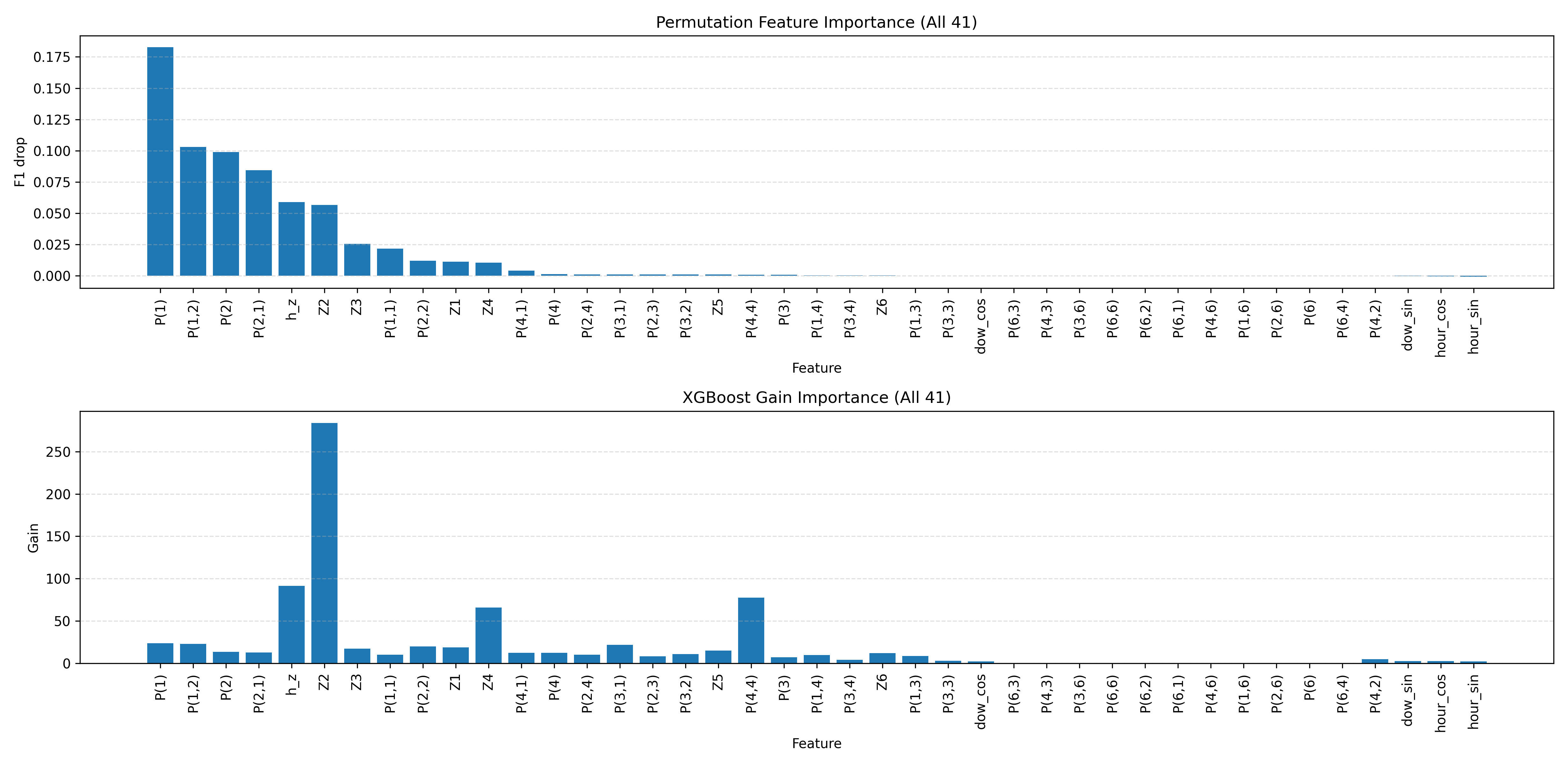}
    \caption{Permutation feature importance and XGBoost gain importance show that P(view), P(view to detail), hz, Z2, and Z3 are important for model performance, measured by the decrease in F1-score. The gain importance indicates that hz, Z2, Z4, and P(add to add) contribute to reducing the training loss.}
    \label{fig:chart}
\end{figure}

\begin{figure}[H]
    \centering
    \includegraphics[width=0.8\textwidth]{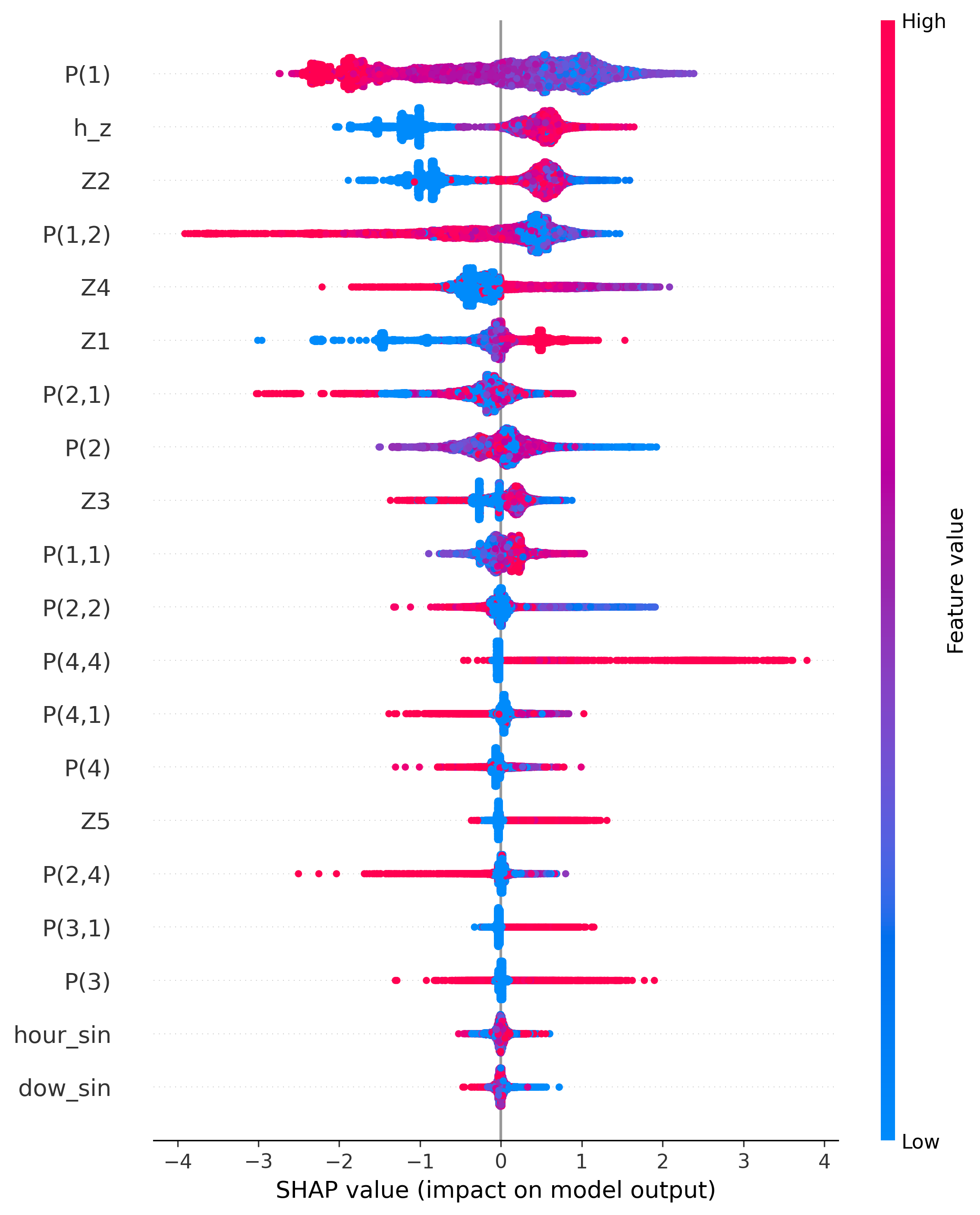}
    \caption{The SHAP summary plot shows the features that matter the most overall. Lower values of P(view) influence class 1 (frustration) and higher values influence class 0 (non-frustration). Similarly, from top to bottom, we can see the feature our XGBoost model relies on the most and its influence on both classes.}
    \label{fig:chart}
\end{figure}

\subsection{LSTM Classifier}\label{subsec2}

The LSTM classifier performs very well. It reaches about 91\% accuracy, F1-score 0.91, and ROC AUC 0.97 on the test set. It has high precision and recall for both classes, with recall around 0.89 for non-frustrated (0) and 0.93 for frustrated (1), so it is good at both catching frustrated sessions and not over-flagging the non-frustrated ones.

Compared to the baseline, XGBoost ( 90\% accuracy, F1 0.90, ROC-AUC 0.96), the LSTM gives a small but consistent improvement in all three metrics (Fig. 5). This suggests that using the full interaction sequence with an LSTM helps the model capture complex patterns that are beyond what our tabular features can provide.

\begin{figure}[H]
    \centering
    \includegraphics[width=0.8\textwidth]{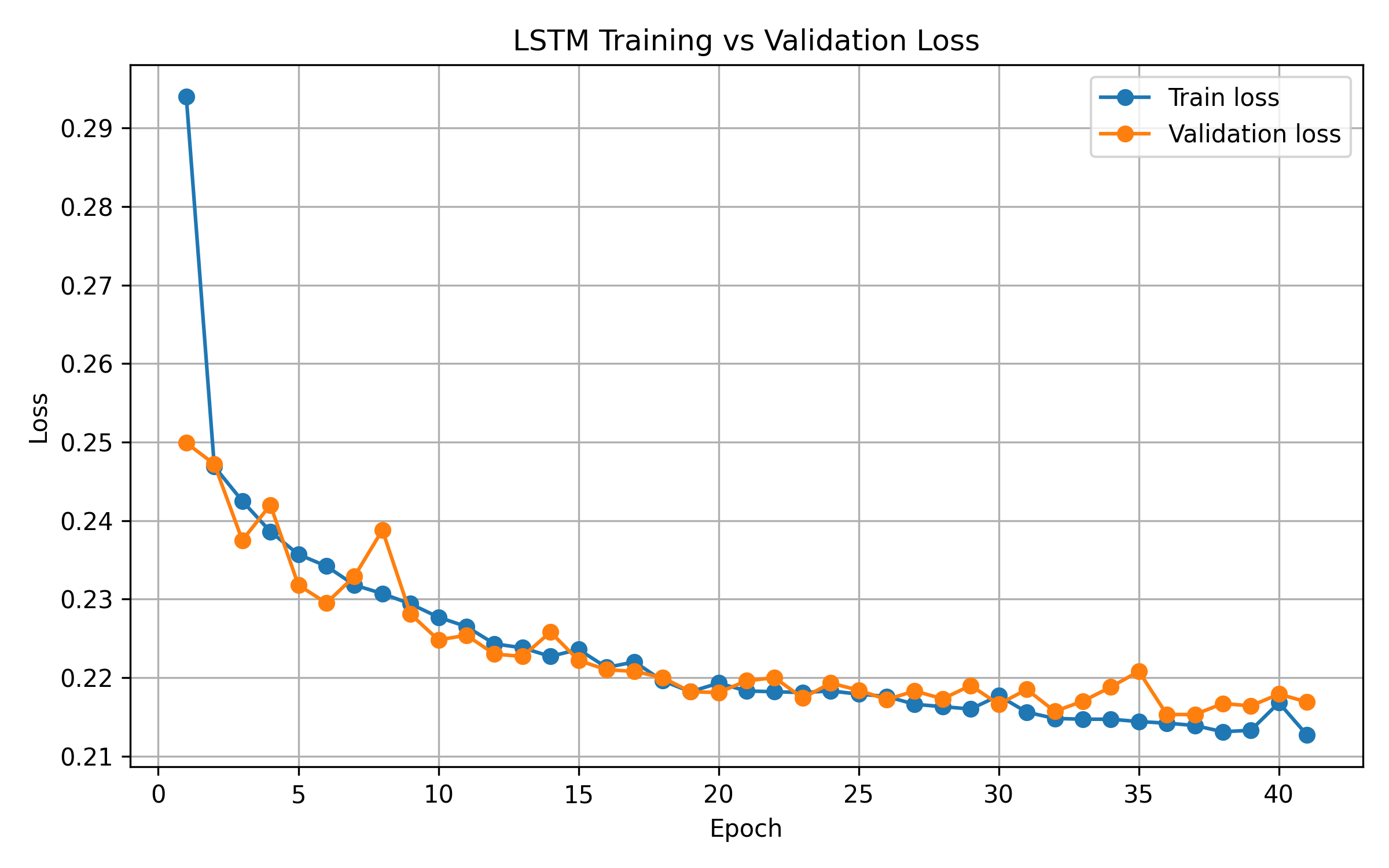}
    \caption{Training vs validation loss curve for the LSTM Classifier shows that it trains and generalizes very well.}
    \label{fig:chart}
\end{figure}

\begin{table}[h!]
\centering

\begin{minipage}{0.47\textwidth}
\centering
\caption{LSTM Classifier: Overall Test Performance}
\begin{tabular}{lc}
\toprule
\textbf{Metric} & \textbf{Value} \\
\midrule
Test F1-score & 0.91139 \\
Test ROC AUC  & 0.97046 \\
Test Accuracy & 0.90970 \\
\bottomrule
\end{tabular}
\label{tab:lstm-overall}
\end{minipage}
\hfill
\begin{minipage}{0.50\textwidth}
\centering
\caption{LSTM Classifier: Detailed Classification Report}
\begin{tabular}{lcccc}
\toprule
\textbf{Class} & \textbf{Precision} & \textbf{Recall} & \textbf{F1-score} & \textbf{Support} \\
\midrule
0 & 0.9261 & 0.8904 & 0.9079 & 8647 \\
1 & 0.8944 & 0.9290 & 0.9114 & 8647 \\
\midrule
\textbf{Accuracy}     & \multicolumn{3}{c}{0.9097} & 17294 \\
\textbf{Macro Avg}    & 0.9103 & 0.9097 & 0.9096 & 17294 \\
\textbf{Weighted Avg} & 0.9103 & 0.9097 & 0.9096 & 17294 \\
\bottomrule
\end{tabular}
\label{tab:lstm-class-report}
\end{minipage}

\end{table}
\vspace{-15pt}

\begin{figure}[H]
    \centering
    \includegraphics[width=0.8\textwidth]{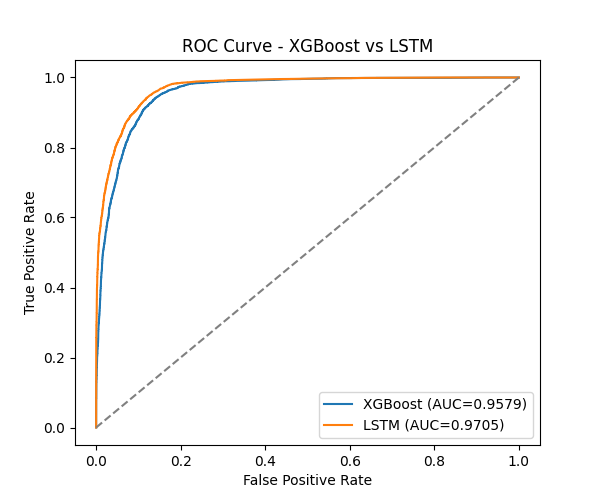}
    \caption{The combined ROC curve shows that both XGBoost and LSTM separate the classes (0, 1) very well, with curves close to the top-left corner. The LSTM curve is slightly above the XGBoost curve (AUC ~0.97 vs ~0.96), meaning that for the same false positive rate, LSTM usually catches a bit more frustrated sessions, so it is the slightly better classifier overall.}
    \label{fig:chart}
\end{figure}

\subsection{Early window prediction}\label{subsec2}
 Here, the idea is to predict whether the session is going to be frustrated or non-frustrated without seeing the full sequence. In the real-time frustration detection use case, we may need to help the frustrated users on the live session, not after the user leaves the digital channel. I use our best model, the LSTM Classifier, for early window prediction. The sequences are truncated to a maximum of 5, 10, 15, 20, and 30 interactions to predict frustration. Sequences shorter than the specified window retain all their available interactions. The objective here is to detect how early in a user’s session we can detect frustration, and that information might be useful to offer help in real-time using automated chatbots, or by making UI updates and improving the digital experience. (Tables 13 and 14)

\textbf{First 5 interactions:}
Performance is almost like random guessing (Accuracy ~0.50, F1 ~0.01). The model predicts almost everything as non-frustrated (class 0 recall ~1.0), and it barely catches any frustrated sessions (class 1 recall ~0.007).

\textbf{First 10 interactions:}
Slight improvement (F1 0.17, ROC AUC 0.76), but the model
is still unable to predict most of the frustration (class 1) and is heavily biased toward class 0.

\textbf{First 15 interactions:}
Now the model starts to become useful (F1 ~0.53, ROC AUC ~0.91). It still favours class 0, but it can already detect a non-trivial fraction of frustrated sessions (class 1 recall ~0.37).

\textbf{First 20 interactions:}
Performance becomes strong (Accuracy ~0.88, F1 ~0.87, ROC AUC ~0.94), and precision/recall for both classes are high and balanced. With about 20 interactions, the model can reliably tell frustrated vs non-frustrated sessions.

\textbf{First 30 interactions:}
Metrics (~0.91 F1, ~0.96 ROC AUC, ~0.91 Accuracy) are very close to the full-sequence LSTM, meaning that most of the useful frustration signal is contained in the first 20–30 interactions of a session.

\begin{table}[h!]
\centering
\caption{Overall Performance for Early-Window Prediction Across Different Interaction Windows}
\begin{tabular}{lccc}
\toprule
\textbf{Interaction Window} & \textbf{F1-score} & \textbf{ROC AUC} & \textbf{Accuracy} \\
\midrule
First 5 events  & 0.0135 & 0.6272 & 0.5032 \\
First 10 events & 0.1729 & 0.7602 & 0.5452 \\
First 15 events & 0.5274 & 0.9053 & 0.6659 \\
First 20 events & 0.8724 & 0.9427 & 0.8751 \\
First 30 events & 0.9069 & 0.9633 & 0.9055 \\
\bottomrule
\end{tabular}
\label{tab:early-window-overall}
\end{table}

\begin{table}[h]
\centering
\caption{Detailed Classification Metrics for Early-Window Prediction}
\begin{tabular}{lcccccc}
\toprule
\multirow{2}{*}{\textbf{Window}} & \multicolumn{3}{c}{\textbf{Class 0}} & \multicolumn{3}{c}{\textbf{Class 1}} \\
\cmidrule(lr){2-4}\cmidrule(lr){5-7}
 & Prec. & Rec. & F1 & Prec. & Rec. & F1 \\
\midrule
5 events  & 0.5016 & 0.9995 & 0.6680 & 0.9365 & 0.0068 & 0.0135 \\
10 events & 0.5238 & 0.9954 & 0.6864 & 0.9536 & 0.0951 & 0.1729 \\
15 events & 0.6046 & 0.9589 & 0.7416 & 0.9008 & 0.3728 & 0.5274 \\
20 events & 0.8597 & 0.8965 & 0.8777 & 0.8919 & 0.8537 & 0.8724 \\
30 events & 0.9177 & 0.8909 & 0.9041 & 0.8940 & 0.9201 & 0.9069 \\
\bottomrule
\end{tabular}
\label{tab:early-window-class-report}
\end{table}

\section{Discussion}\label{sec2}

\subsection{Key findings linked to objectives}\label{subsec2}

\textbf{1 - Define user frustration:}
This research showed that using rage clicks, back-and-forth navigation (U-turns), cart churn, search struggle, and long wandering sessions, we can define digital frustration from sessionized clickstream data. Using the combination of rules, I was able to label about 19\% of the 304,881 sessions as frustrated. Each type of label can be explained by a clear user behaviour pattern based on the rules mentioned in section 3.2.

\textbf{2 - Labeling strategy and features:}
From 5.4M raw clickstream events downloaded from Coveo \cite{29}, I built a pipeline that creates session-level frustration labels and engineers new features. The features include 1-gram and 2-gram frequencies, HVG motifs, motif entropy, and cyclical features. These engineered features work well for our standard models as predictors, and feature importance/SHAP results (Fig. 2, Fig. 3) show that a set of n-gram and motif features is more influential for frustration detection.

\textbf{3 - Standard  vs deep learning models:}
Among the standard machine learning models, logistic regression reaches 84\% accuracy, Random Forest 89\%, and XGBoost 90\% (ROC AUC up to 0.96). The deep learning based sequence model (LSTM Classifier) was the best one (with 91\% accuracy, F1 0.91, and ROC AUC 0.97). The results show that the standard models with engineered features as input give us a very good baseline, and the LSTM model with only the interaction sequence as input gives a small but consistent improvement from the baseline.

\textbf{4 - How early can we predict?:}
When we feed in the first 5–10 interactions, the LSTM is close to only a random guess and rarely detects frustration. Using the first 15 interactions, it becomes relatively useful. From 20 interactions onward, the performance of the model becomes strong (F1 0.87, ROC AUC 0.94), and by using the first 30 interactions, the results become almost as good as using the full interaction sequence. This means we can start making reliable frustration predictions once a user has made around 20–30 interactions in the digital channel.

\textbf{5 - How do the adopted features contribute to digital frustration detection?} - As we discussed in section 4.5, I adopt the same family of 1-gram, 2-gram, and HVG motif features \cite{16}. The experiment shows that these features are very good predictors for the standard classifiers when it comes to digital frustration.

\subsection{Limitations and Further Improvements}\label{subsec2}

In this research, I focused on three standard classifier models and one LSTM Classifier architecture. There are transformer-based approaches available (e.g., TRACE \cite{43}), and those are not used.  For early-window prediction, I did not create features or train separate models on the sliced interaction sequences. I reused the same LSTM Classifier trained on the full sequences (length between 2 and 1000) on the
sliced sequences. Dedicated early-window models trained only on the first N events, and features could further improve early frustration detection.

Another important limitation is that I'm building the frustration detection models on a balanced dataset (50\% frustration and 50\% non-frustration). The models are not trained, and no results are reported on the original imbalanced data (18.91\% frustration and 81.09\% non-frustration)

\section{Conclusion}\label{sec2}

In this research, I showed that it is possible to detect digital frustration from raw clickstream data using machine learning, using the full sequence as well as an early window. I defined a rule-based frustration label (based on rage bursts, U-turns, cart churn, search struggle, and long wandering sessions), built a reproducible pipeline that processes 5.4M raw interactions into 304,881 labelled sessions, and engineered tabular features that can be used by standard machine learning models and developed a discriminative LSTM classifier for sequence modeling.

The results show that standard models like XGBoost achieve about 90\% accuracy, ROC AUC of 0.9579, and a discriminative LSTM classifier using the full event sequence gives a small but consistent improvement, reaching about 91\% accuracy and a ROC AUC of 0.9705. I also studied early-window prediction and found that the most useful predictive signal appears within the first 20-30 interactions, where the LSTM can already predict frustration reliably.

\section*{Declarations}

\subsection*{Funding}
No funding was received to assist with the preparation of this manuscript.

\subsection*{Competing interests}
The author has no relevant financial or non-financial interests to disclose.

\subsection*{Data availability}
The data used in this study are publicly available through the Coveo e-commerce dataset \cite{29}.

\subsection*{Code availability}
The code is publicly available at https://github.com/jibin10/clickstream-frustration-prediction.

\bigskip





\bibliography{sn-bibliography}% common bib file

\end{document}